\documentclass[runningheads]{llncs}
\usepackage{graphicx}
\usepackage{array}
\usepackage{booktabs}

\usepackage[figuresright]{rotating}
\usepackage{xcolor}
\usepackage{float}
\usepackage{longtable}
\usepackage{setspace}
\usepackage{hyperref}
\usepackage{tikz}
\usepackage{caption} 
\usepackage{longtable}

\begin{document}
\title{ConceptLens: from Pixels to Understanding} 
\titlerunning{ConceptLens: from Pixels to Understanding}
%
\author{Abhilekha Dalal\inst{1} \and
Pascal Hitzler\inst{1}}
\authorrunning{Abhilekha Dalal, Pascal Hitzler}
%
\institute{Kansas State University, Manhattan, KS, USA} 
\maketitle              
\setlength{\tabcolsep}{6pt}
\begin{abstract}

ConceptLens is an innovative tool designed to illuminate the intricate workings of deep neural networks (DNNs) by visualizing hidden neuron activations. By integrating deep learning with symbolic methods, ConceptLens offers users a unique way to understand what triggers neuron activations and how they respond to various stimuli. The tool uses error-margin analysis to provide insights into the confidence levels of neuron activations, thereby enhancing the interpretability of DNNs. This paper presents an overview of ConceptLens, its implementation, and its application in real-time visualization of neuron activations and error margins through bar charts.

\end{abstract}

\section{Introduction}

Deep neural networks (DNNs) have revolutionized various fields by offering powerful solutions for complex tasks such as image recognition, natural language processing, and more~\cite{ramprasath2018image,graves2014towards,auli2013joint}. However, the "black box" nature of these models often leaves users and researchers in the dark about how specific decisions are made~\cite{apple_card,CEO_bias}. Understanding the internal workings of these networks is crucial for improving their reliability and trustworthiness. 
A promising approach to make them more interpretable is to associate the activations of neurons in their hidden layers with human-understandable concepts \cite{dalal2024value,oikarinen2022clip,ace_ghorbani}. Prior work \cite{dalal2024value} has focused on identifying the concepts that maximally activate each neuron -- corresponding to the notion of recall. However, solely optimizing for recall is insufficient, as neurons tend to also activate for many other inputs that do not match their assigned concept (low precision).

To address this limitation, we present a visualizing tool, \textit{ConceptLens}, that quantifies the uncertainty and imprecision in neural concept labels through error margins. \textit{ConceptLens} leverages the principles outlined in the research paper~\cite{dalal2024error}, which uses symbolic Semantic Web methods to automatically induce semantic concept labels for individual neurons from a large knowledge base made from Wikipedia categories and evaluate their precision by analyzing the false positive rates of neuron activations. This approach allows users to see not only what stimuli activate specific neurons but also how confidently these neurons respond to different inputs.
\section{Method}
\label{sec:method}

\paragraph{\textbf{System Overview}} The core idea behind \textit{ConceptLens} is to provide bar chart visualizations that contextualize the certainty of each detected concept based on the neuron activations. \textit{ConceptLens} combines a Convolutional Neural Network (CNN) trained on specific image classes with symbolic reasoning techniques (Concept Induction) to assign semantically meaningful labels to the neurons in the final dense layer from a knowledge base of 2 million concepts. The system's backend processes images to detect concepts and calculate error-margin percentages, indicating the confidence level of each activation.

\paragraph{\textbf{Error-Margin Analysis}} The core innovation of ConceptLens lies in its error-margin analysis. This measure assesses the likelihood that a given neuron activation accurately corresponds to the assigned concept by evaluating how frequently neurons activate for concepts not assigned to them on a holdout set of images. Lower error-margin percentages indicate higher confidence, while higher percentages suggest greater uncertainty. This dual focus on recall (identifying activating stimuli) and precision (evaluating responses to non-target stimuli) provides a comprehensive understanding of neuron behavior.

\paragraph{\textbf{User Interface}} \textit{ConceptLens} features a user-friendly interface that allows users to upload images and receive real-time visualizations of neuron activations. The main components of the interface include:

\begin{enumerate}
    \item \textbf{Image Upload and Selection:} Users can upload their images or choose from a curated gallery. The tool supports a wide range of images, although results may vary for images outside the 10 classes it was primarily trained on: bathroom, bedroom, building facade, conference room, dining room, highway, kitchen, living room, skyscraper, and street.
    \item \textbf{Concept Detection and Visualization:} ConceptLens processes the uploaded image through trained CNN and Concept Induction to detect relevant concepts. The detected concepts are then presented as bar chart visualization and their corresponding error-margin percentages, providing users with a clear understanding of the network's predictions.
    \item \textbf{Error-Margin Display:} The interface highlights the error-margin percentages for each detected concept, allowing users to gauge the confidence of the network's predictions. Lower percentages indicate higher confidence in the concept detection.
\end{enumerate}

\paragraph{\textbf{Technical Details}}
\textit{ConceptLens} utilizes a ResNet50V2 architecture for its CNN, trained on a subset of the ADE20K dataset. The network's last hidden layer neurons are analyzed (see 
\cite{dalal2024value}) and labeled using an OWL-reasoning-based Concept Induction algorithm (ECII, 
\cite{conceptinduction_sarker}) over a large background knowledge base derived from Wikipedia \cite{wikikg_sarker}.  This assigns high-level concepts to neurons, facilitating the error-margin analysis.

Error margins are calculated by evaluating neuron activations across a large dataset of images from Google and ADE20K (see \cite{dalal2024error}). This includes both target label images (those that match the neuron’s assigned concept) and non-target label images (those that do not match the concept). By comparing activation patterns, \textit{ConceptLens} determines the likelihood of correct concept detection, thus providing valuable insights into the network's interpretability.

\begin{figure}
    \centering
    \includegraphics[width=\textwidth, height=10cm]{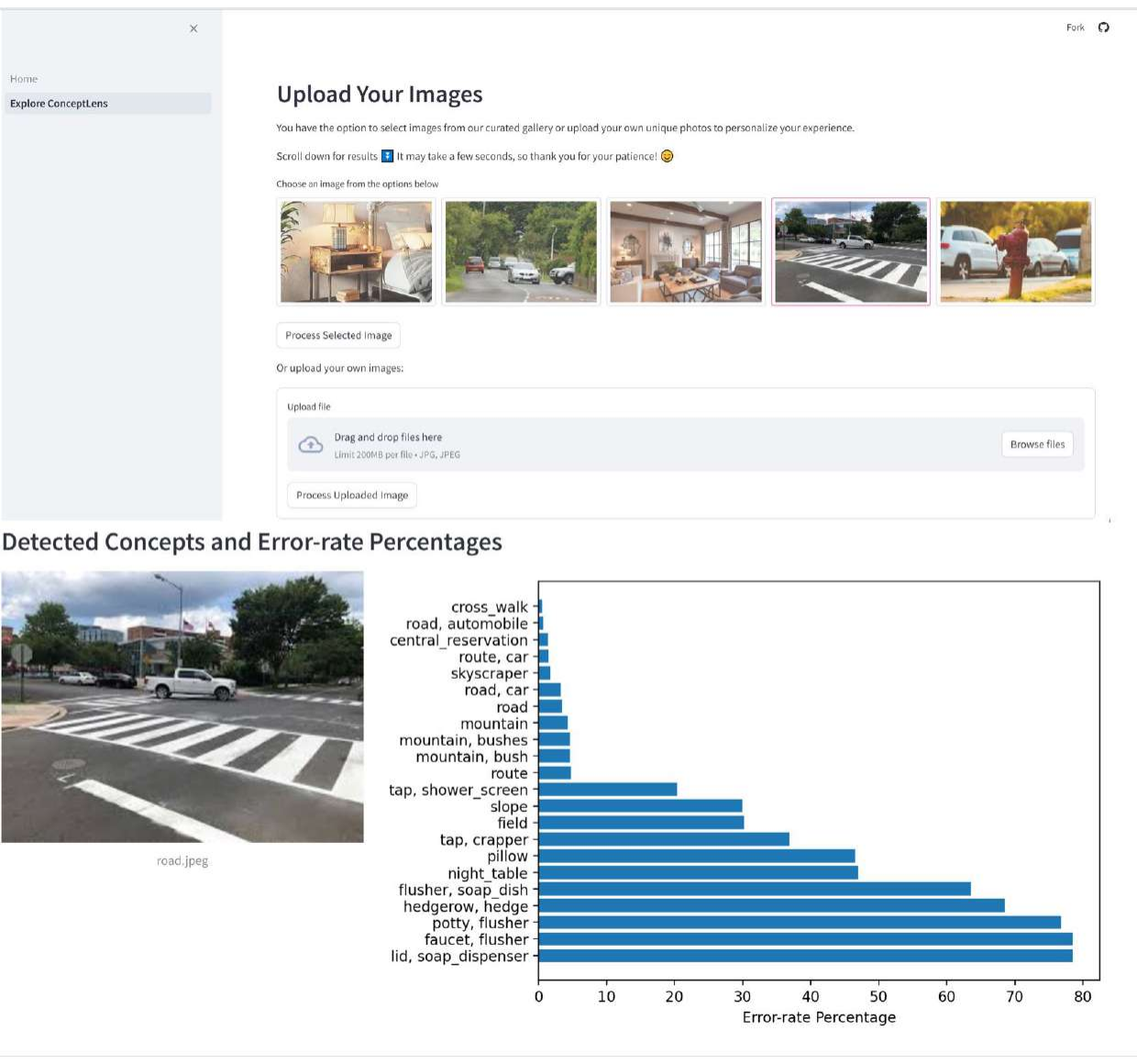}
    \caption{ConceptLens visualizes detected concepts in images, showing their error margins. In this street scene, "cross\_walk" and "road" are confidently recognized with low error percentages, while uncertainty is shown for other labels like "automobile" and "central\_reservation".}
    \label{fig:view}
\end{figure}

\paragraph{\textbf{Demonstration}}
Explore ConceptLens firsthand through our interactive tool at: ConceptLens Demo.\footnote{\url{https://conceptlens.streamlit.app/Explore_ConceptLens}} Watch the demo video for a preview of its features here\footnote{\url{https://youtu.be/yLYig1IjB9Y}} and find the code repository on GitHub\footnote{\url{https://github.com/abhilekha-dalal/ConceptLens}} for deeper exploration and implementation.

\section{Conclusion}
\label{sec:conclusion}

ConceptLens represents a pioneering advancement in explainable AI, offering a robust tool for visualizing and interpreting hidden neuron activations within neural networks. By integrating advanced error-margin analysis with convolutional neural networks and symbolic reasoning techniques, ConceptLens bridges critical gaps in model interpretability. Key areas for future development include: 1) Extending ConceptLens to a broader range of datasets and classes 
2) Improving the user interface based on continuous user feedback 
3) Developing more sophisticated error-margin analysis methodologies for deeper insights into neural network reliability.

\textbf{Acknowledgement} Authors acknowledge partial funding under NSF grant 2333532 \emph{EduGate}.


\bibliographystyle{splncs04}
\bibliography{reference}
\newpage

\end{document}